\documentclass[conference]{IEEEconf}
\IEEEoverridecommandlockouts

\usepackage{siunitx}
\usepackage{physics}
\usepackage[T1]{fontenc}
\AtBeginDocument{\RenewCommandCopy\qty\SI}
\usepackage{amsmath,amsfonts,amssymb}
\usepackage{array}
\usepackage{cite}
\usepackage[caption=false,font=scriptsize]{subfig}
\usepackage{textcomp,stfloats,float,url,verbatim,graphicx}
\usepackage{pifont,algorithm,algpseudocode,hyperref,cleveref,comment}
\usepackage{multirow,booktabs,rotating,makecell,tabularx,xcolor}
\usepackage{tikz}
\usepackage{caption}
\usetikzlibrary{positioning,shapes,arrows,calc}

\usepackage{xcolor}

\definecolor{TUMBlue}{cmyk}{1,0.43,0,0}
\definecolor{TUMWhite}{cmyk}{0,0,0,0}%
\definecolor{TUMBlack}{cmyk}{0,0,0,1}%
\definecolor{TUMBlue1}{cmyk}{1,0.57,0.12,0.7}
\definecolor{TUMBlue2}{cmyk}{1,0.54,0.04,0.19}
\definecolor{TUMGray1}{cmyk}{0,0,0,0.8}%
\definecolor{TUMGray2}{cmyk}{0,0,0,0.5}%
\definecolor{TUMGray3}{cmyk}{0,0,0,0.2}%
\definecolor{TUMBlue3}{cmyk}{0.65,0.19,0.01,0.04}
\definecolor{TUMBlue4}{cmyk}{0.42,0.09,0,0}
\definecolor{TUMIvory}{cmyk}{0.03,0.04,0.14,0.08}%
\definecolor{TUMOrange}{cmyk}{0,0.65,0.95,0}%
\definecolor{TUMGreen}{cmyk}{0.35,0,1,0.2}%

\def\BibTeX{{\rm B\kern-.05em{\sc i\kern-.025em b}\kern-.08em
    T\kern-.1667em\lower.7ex\hbox{E}\kern-.125emX}}

\begin{document}

\title{\LARGE\bf
Drifting in the Future: Stabilizing Path Following Drifting on High-Latency Vehicle Systems}

\author{Frederik Werner, Till Heintzenberg, Markus Lienkamp, Johannes Betz
\thanks{F. Werner, J. Betz are with the Professorship of Autonomous Vehicle Systems, TUM School of Engineering and Design, Technical University of Munich, 85748 Garching, Germany; Munich Institute of Robotics and Machine Intelligence (MIRMI), corresponding author: frederik.werner@tum.de
This work has been accepted for publication in the Proceedings of the IEEE International Conference on Robotics and Automation (ICRA 2026) IEEE.}}

\maketitle
\begin{abstract}
Autonomously controlling and handling a vehicle at and beyond its stability limit is a mathematically and computationally demanding task. Prior demonstrations of automated drifting have been limited to research platforms with instantaneous torque delivery and independently actuated wheels, leaving their applicability to production vehicles with actuator latencies and mechanically coupled axles uncertain. To overcome these issues, we design a predictor to compensate for powertrain delays, develop a revised control formulation to accommodate higher actuation latencies as well as a differential coupling on the driven axle, and introduce brake-based velocity stabilization. This paper presents the controller framework, the model extensions, and real-world experimental results. We observe that our controller enables a production sports car with a combustion engine to robustly sustain circular and figure-eight drifts, limiting lateral error to 1.1 m and sideslip overshoot to 0.06 rad despite actuator delays exceeding 250 ms, while mitigating oscillations and maintaining stable path and sideslip tracking.
In conclusion, our results establish that autonomous drifting is feasible on production-ready vehicles, opening pathways to advanced safety systems capable of stabilizing cars in scenarios where traditional control fails.
\end{abstract}

\section{Introduction}

Conventional safety systems like Electronic Stability Control (ESC) ensure autonomous vehicles operate within stable handling limits. These systems are designed to prevent excursions into open-loop unstable regimes, ensuring controllability for the average driver. However, to surpass expert human drivers in safety and performance, autonomous vehicles must master the full range of vehicle dynamics, especially in unexpected situations \cite{Betz.2022}. Emergency maneuvers or sudden changes in road friction may push the vehicle beyond traditional stability boundaries, a domain where expert drivers use drifting techniques for control \cite{Abdulrahim2006}.

While earlier research \cite{Olofsson.2013, Joa.2020, Yin.2020, Weber.2024b, Liu.2025} has demonstrated the viability of autonomous drifting along a path, its validation was performed on a purpose-built electric research vehicle with near-instantaneous torque delivery and independently actuated rear wheels \cite{Goh.2020}. We found that direct application of this framework to a production vehicle is not possible without adaptations. Vehicles equipped with an internal combustion engine (ICE) present a more complex problem due to actuator delays in the powertrain and mechanical coupling of the rear wheels through a differential, which prohibits independent wheel speed control.
This work addresses these challenges by adapting and extending a validated control framework for use on a high-performance, ICE-powered production sports car. We demonstrate our novel controller on a real-world vehicle and show that we can successfully initiate, sustain, and transition between drifts while accurately tracking dynamic paths, including circles and figure-eight maneuvers. In doing so, this research takes a step toward transferring drift control from research platforms to real-world production vehicles, supporting its potential for future safety and performance systems.

\begin{figure}
\centering
\includegraphics[width=1\columnwidth]{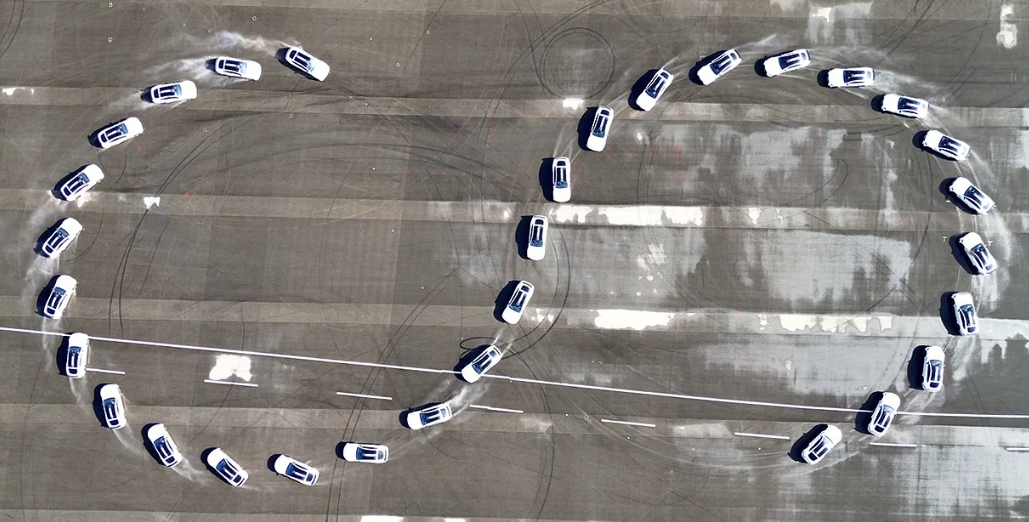}
\caption{Demonstration of our autonomous drifting control algorithm in a production vehicle performing figure-eight drifting on a surface with mixed wet, damp, and dry conditions. \cite{Boskovic.2024}}
\label{fig:figure-eight}
\end{figure}

This work extends a previously validated drift control framework \cite{Goh.2020} to production-ready high-latency vehicles, introducing several novel elements that make autonomous drifting feasible on a conventional ICE sports car. Our main contributions are:
\begin{itemize}
    \item Design of a state prediction module that compensates for ICE powertrain delays and phase-aligns the steering and torque dynamics, preventing oscillations.
    \item Development of a revised drift control formulation for a vehicle with high actuation latencies and without independent wheel-torque actuation.
    \item Integration of brake-based velocity stabilization to counter overshoots during drift transitions and improve path tracking accuracy.
\end{itemize}

Videos of our experiments are available at: \url{https://github.com/TUM-AVS/ICRA2026_Drifting_in_the_future}


\section{RELATED WORK}

Research on autonomous drifting \cite{Liu.2025} initially focused on open-loop or equilibrium stabilization using nested control structures \cite{Voser.2010, Hindiyeh.2014, VELENIS20111363, Gonzales2016, Acosta2018}. These methods demonstrated drift stabilization at fixed points but lacked path-following capabilities.

Bárdos et al. \cite{Bardos.2020} proposed a linear-quadratic regulator around a known equilibrium, but their approach did not generalize to varying curvature. Similarly, the model-free strategy by Joa et al. \cite{Joa.2020} achieved equilibrium control using only measurable signals, omitting tire models and friction estimation.

Goh and Gerdes \cite{Goh.2020} developed a control architecture capable of maintaining a drift while performing path-following. Their framework imposed desired dynamics on lateral and sideslip errors and used model inversion to compute control inputs. Later works by Goh \textit{et al.} included a nonlinear model predictive controller (MPC) for racetrack drifting \cite{Goh.2024} and the definition of recoverability envelopes \cite{Goh.2019}, as well as the use of brakes to tackle the under-actuation problem for the stabilization of drifts along a path \cite{Goel.2020}. His work on MPC showed the capability to handle ICE drivetrains as well, but drawbacks include high computational requirements and sensitivity to model mismatch.
In addition, multiple authors focused on MPC variations for autonomous drifting \cite{Dong2022, Stano2024} but demonstrated this only in simulations.
Recently, Meyjer \textit{et al.} developed an MPC based controller for a production BMW M3 vehicle, but only as a driver aid, not in a fully autonomous setup \cite{meijer2024nonlinearmodelpredictivecontrol}.
Djeumou \textit{et al.} investigated drifting in both specialized drift vehicles and production vehicles using an MPC based on a physics-informed conditional diffusion model
\cite{djeumou2024one, GohTireModels}. Their approach achieved remarkable improvements in generalization capability while requiring comparatively little training data. However, reliance on a high-performance CPU for execution indicates substantial computational demands.

Learning-based approaches \cite{Yin.2020, Tth2024, Jiang.2021c, zhou2025adaptive, zhou2025learning, Cai.2020, DeepDrifting2022} have also been explored, where agents are trained in simulation environments to acquire drifting skills. These controllers show promising flexibility and adaptability, being able to discover non-intuitive control strategies and operate without requiring detailed vehicle or tire models.  However, most published results remain confined to simulation studies, with little evidence of reliable real-world deployment due to challenges in sim-to-real transfer, robustness to model mismatch, and the need for extensive training data. 

\subsection{Baseline Controller}
\label{subsec:baseline}
 Goh \textit{et al.} ~\cite{Goh.2016, Goh.2020} introduced a control framework for autonomous drifting along trajectories. This work was validated on the MARTY DeLorean test vehicle with independently actuated, electrically powered rear wheels. Their research demonstrated robust sideslip and path tracking during figure-eight drifts with a maximum lateral error of \SI{1.5}{m}, a maximum heading error of \SI{0.1}{\radian} and a maximum sideslip overshoot of \SI{5}{\degree} \cite{Goh.2019}. In addition, the framework is computationally efficient, which makes it a promising candidate for adaptation to production-ready software.  We are building on the control framework of Goh \textit{et al.} ~\cite{Goh.2016, Goh.2020} for path and sideslip tracking which consists of four key modules (Fig.~\ref{fig:InternalControllerStructure}):

\begin{figure}[h]
    \centering
    \begin{tikzpicture}[rotate=0, transform shape, >=stealth', thick, node distance=4cm]

\tikzstyle{block} = [draw, rectangle, minimum height=2em, minimum width=3em]

\node [block, minimum height=4.2em, text width = 5.5em, TUMBlack] (IED) {Imposed Error\\Dynamics};
\node [block, below of=IED, node distance=2.7cm, minimum height=7em, text width = 4.5em, xshift=-5mm, TUMBlue] (NMI) {Nonlinear Model Inversion};
\node [block, right of=NMI, node distance=3cm, minimum height=4em, text width = 4em, xshift=12mm, TUMGreen] (TSM) {Tire Slip Mapping};
\node [block, above of=TSM, node distance=2.8cm,minimum height=4.5em, text width = 5.5em, xshift=-4mm, TUMOrange] (WSL) {Wheel Speed Control Loop};

\draw [->] ([xshift=-18mm,yshift = 6mm]IED.west) -- ([yshift = 6mm]IED.west) node[midway, above] {$r_{\text{ref}},\beta_{\text{ref}},\Dot{\phi}_{\text{ref}}$};
\draw [->] ([xshift=-18mm,yshift = 0mm]IED.west) -- ([yshift = 0mm]IED.west) node[midway, above] {$r,\beta,V$};
\draw [->] ([xshift=-18mm,yshift = -6mm]IED.west) -- ([yshift = -6mm]IED.west) node[midway, above] {$e,\Delta\phi$};

\draw [->, TUMBlack] ([xshift=-5mm]IED.south) -- node[right] {$\Dot{r}_{\text{des}},\Dot{\phi}_{\text{des}}$} ([yshift = 0mm]NMI.north);
\draw [->] ([xshift=-12mm,yshift = 3mm]NMI.west) -- ([yshift = 3mm]NMI.west) node[midway, above] {$r,\beta,V$};
\draw [->] ([xshift=-12mm,yshift = -3mm]NMI.west) -- ([yshift = -3mm]NMI.west) node[midway, above] {$\omega$};
\draw [->] ([xshift=-12mm,yshift = -9mm]NMI.west) -- ([yshift = -9mm]NMI.west) node[midway, above] {$a_x,a_y$};

\draw [->, TUMBlue] ([yshift = -10mm]NMI.east) -- ([xshift=51mm,yshift = -10mm]NMI.east) node[midway, below] {$\delta_{\text{cmd}}$};

\draw [->, TUMBlue] ([yshift = 3mm]NMI.east) -- ([yshift = 3mm]TSM.west) node[midway, above] {$\gamma_{\text{des}}$};

\draw [->, TUMBlue] ([yshift = 10mm]NMI.east) -- ([xshift = 24mm, yshift = 10mm]NMI.east) node[above, midway] {$F_{xr,des}$}  -| ([xshift = -4mm]WSL.south);

\draw [->] ([xshift=-15mm,yshift = -3mm]TSM.west) -- ([yshift = -3mm]TSM.west) node[midway, above] {$r,\beta,V$};

\draw [->, TUMGreen] ([yshift = 0mm]TSM.north) -- ([xshift=4mm,yshift = 0mm]WSL.south) node[midway, right] {$\omega_{\text{des}}$};

\draw [->] ([xshift=-8mm,yshift = 0mm]WSL.west) -- ([yshift = 0mm]WSL.west) node[midway, above] {$\omega$};


\draw [->, TUMOrange] ([yshift = 0mm]WSL.east) -- ([xshift = 12mm] WSL.east) node[midway, above] {$\tau_{\text{cmd}}$};

\end{tikzpicture}
    \caption{Internal structure of the drift controller adapted from Goh \textit{et al.}~\cite{Goh.2020}. The controller consists of four main modules explained in Sec.~\ref{subsec:baseline} and adapted in Sec.~\ref{subsec:adaptations}.}
    \label{fig:InternalControllerStructure}
\end{figure}
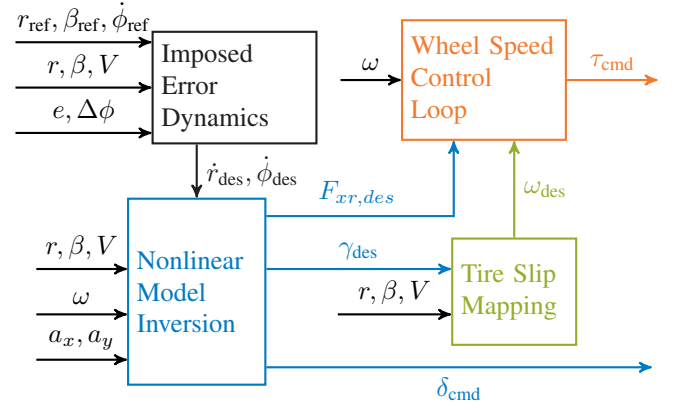

\subsubsection{Imposed Error Dynamics} Stable dynamics are imposed on the lateral path error and the sideslip tracking error, yielding a desired course angle rate $\dot{\phi}_{des}$ and yaw acceleration $\dot{r}_{des}$.

\subsubsection{Nonlinear Model Inversion}
This module inverts a single-track vehicle model to calculate the steering angle $\delta_{cmd}$ and rear tire thrust angle $\gamma_{des}$ required to produce the desired yaw and course angle rates ($\dot{r}_{des}$, $\dot{\phi}_{des}$). To handle actuator limits, this calculation is constrained to a feasible set of achievable dynamics, labeled as the \textit{tangent space}. When a desired state derivative is unachievable, the \textit{yaw stability preserving projection} (YSPP) prioritizes sideslip tracking over path tracking to prevent the vehicle from losing control authority and spinning out \cite{Goh.2019}.

A key model assumption described by Eq. \ref{eq:FrictionCircle} considers continuously saturated rear wheels. Because of this assumption and given a constant friction value $\mu_r$, the magnitude of the rear tire force is constant and can be represented by the rear thrust angle $\gamma$, which is the angle of the rear tire force vector. 
\begin{align}
    F_{xr} &= \sqrt{\mu_r^2 \, F_{zr}^2 - F_{yr}^2} \label{eq:FrictionCircle}
\end{align}

\subsubsection{Tire Slip Mapping} The Tire Slip Mapping Module converts the rear thrust angle into the required wheel speed $\omega_{des}$.

\subsubsection{Wheel Speed Control Loop} Torque commands $\tau_{cmd}$ are generated to track the desired wheel speed, taking into account wheel inertia and tire forces. Factoring in wheel and drivetrain inertia accelerates the tracking of the desired rear longitudinal force $F_{xr,des}$. Experiments by Goh \textit{et al.}  showed that wheel speed control reduces yaw oscillations and improves path tracking compared to direct calculation of $\tau_{cmd} = R F_{xr,des}$ via the tire rolling radius $R$ \cite{Goh.2020}.

\section{METHOD}

\begin{figure}[h]
\centering
\includegraphics[width=1\columnwidth]{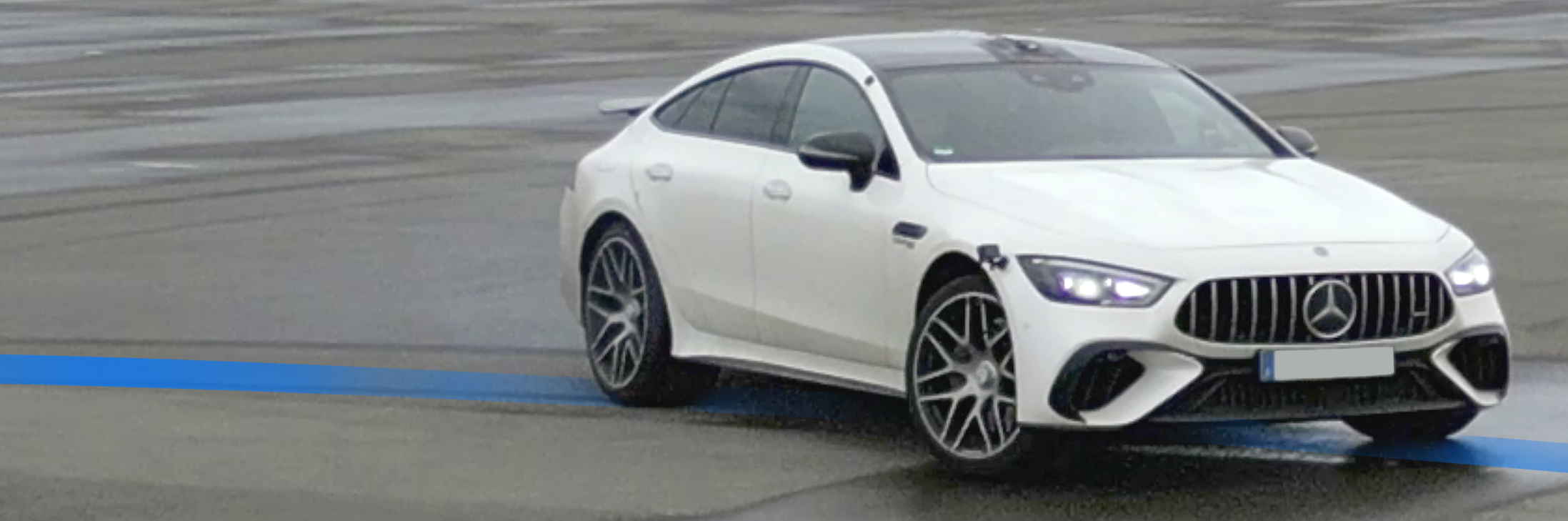}
\caption{The Mercedes-AMG GT\,63\,S research vehicle during a drift maneuver. \cite{Boskovic.2024}}
\label{fig:x290}
\end{figure}

Our target platform is a Mercedes-AMG GT\,63\,S 4-Door shown in Fig.~\ref{fig:x290}, equipped with production actuators that accept torque and steering angle commands via the CAN interface, enabling autonomous operation. The vehicle state is estimated using an Oxford OXTS RT3000 differential INS unit operating at $100\text{ Hz}$.

Compared to the MARTY platform, stabilizing drifts on the AMG poses additional challenges. The primary challenges are the actuator latency of the ICE powertrain and the coupled rear wheels, which prevent independent control of the wheel speed. The following sections outline our adaptations to address these limitations.

\subsection{Adaptations to the Baseline Controller and Integration into a Closed-Loop Simulation}
\label{subsec:adaptations}
We integrated the baseline controller into our closed-loop simulation framework comprising four main modules, shown in Fig.~\ref{fig:GeneralControllerStructure}:

\begin{figure}[h]
    \centering
    \begin{tikzpicture}[rotate=0, transform shape, >=stealth', thick, node distance=4cm]

\tikzstyle{block} = [draw, rectangle, minimum height=2em, minimum width=3em]

\node [block, minimum height=7em, text width = 4.7em, TUMOrange] (driftcontroller) {Drift/ Transitionary Controller};
\node [block, right of=driftcontroller, text width = 4em, node distance=3.4cm, minimum height=7em, TUMBlue] (vehicledynamics) {Vehicle Dynamics};
\node [block, right of=vehicledynamics, node distance=2.8cm, minimum height=9em, yshift = 3mm, TUMGreen] (predictor) {Predictor};
\node [block, above of=predictor, node distance=2.5cm, text width = 4.5em,minimum height=4em,xshift=-30mm, TUMBlack] (trajectorydynamics) {Trajectory Dynamics};

\draw [->, TUMOrange] (driftcontroller) -- node[below] {$\delta_{\text{cmd}}, \tau_{\text{cmd}}$} (vehicledynamics);

\coordinate (midarrow) at ($(driftcontroller.east)!0.5!(vehicledynamics.west)$);

\draw [->, TUMOrange] (midarrow) -- ++(0,1.5) |- node[near end, above] {} ([yshift=12mm]predictor.west);

\draw [->, TUMBlue] ([yshift=9mm]vehicledynamics.east) -- node[above] {$x, y, \phi$} ([yshift=6mm]predictor.west);

\draw [->, TUMBlue] ([yshift=3mm]vehicledynamics.east) -- node[above] {$r, \beta, V$} ([yshift=0mm]predictor.west);

\draw [->, TUMBlue] ([yshift=-3mm]vehicledynamics.east) -- node[above] {$\omega$} ([yshift=-6mm]predictor.west);

\draw [->, TUMBlue] ([yshift=-9mm]vehicledynamics.east) -- node[above] {$\delta,\tau$} ([yshift=-12mm]predictor.west);

\draw [->, TUMGreen] ([xshift=6mm]predictor.north) |- node[above, xshift=-8mm] {$x_P, y_P, \phi_P$} ([yshift=3mm]trajectorydynamics.east);

\draw [->, TUMGreen] ([xshift=0mm]predictor.north) |- node[above, xshift=-4mm] {$r_P,\beta_P,V_P$}([yshift=-3mm]trajectorydynamics.east);

\draw [->, dashed, TUMBlack] ([yshift=-5mm]trajectorydynamics.west)  -| node[near end, right] {s} ([xshift=4mm]driftcontroller.north);

\draw [->, TUMBlack] ([yshift=0mm]trajectorydynamics.west) -- node[above, xshift=-4mm] {$e, \Delta\phi$} ++(-1.5,0) -| ([xshift=0mm,yshift=9mm]driftcontroller.north) -| ([xshift=0mm]driftcontroller.north);

\draw [->, TUMBlack] ([yshift=5mm]trajectorydynamics.west) -- node[above, xshift=-4mm] {$r_{\text{ref}},\beta_{\text{ref}},\Dot{\phi}_{\text{ref}}$} ++(-2,0)  -| ([xshift=-4mm,yshift=9mm]driftcontroller.north) -| ([xshift=-4mm]driftcontroller.north);

\draw [->, TUMGreen] ([xshift=-4mm]predictor.south) |- 
([xshift=-4mm,yshift=-5mm]predictor.south) --
node[above] {$\omega_P$}([xshift=4mm,yshift=-6mm]driftcontroller.south) |-  ([xshift=4mm]driftcontroller.south);

\draw [->, TUMGreen] ([xshift=4mm]predictor.south) |- 
([xshift=4mm,yshift=-11mm]predictor.south) --
node[above] {$r_P,\beta_P,V_P$}([xshift=-4mm,yshift=-12mm]driftcontroller.south) |-  ([xshift=-4mm]driftcontroller.south);

\coordinate (midarrow2) at ($(predictor.east)!0.5!([yshift=-3mm]trajectorydynamics.west)$);

\end{tikzpicture}
    \caption{Overview of the closed-loop control and simulation architecture.}
    \label{fig:GeneralControllerStructure}
\end{figure}
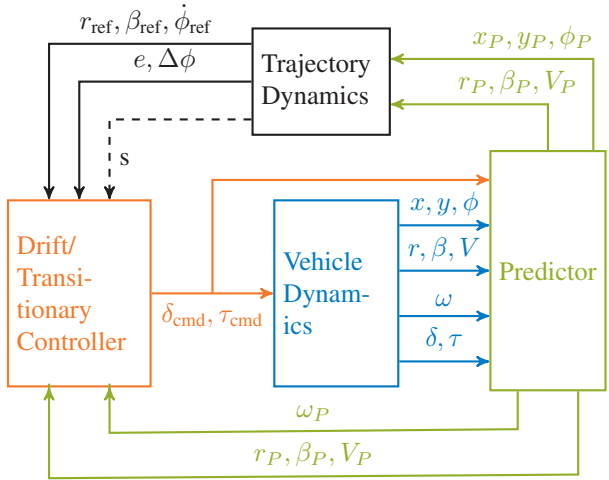

\begin{itemize}
    \item \textbf{Drift/Transitionary Controller:} Contains the described drift control algorithm as well as a simple, Stanley-style controller for path tracking in transitionary phases like starting from a standstill or return-to-start.
    \item \textbf{Vehicle Dynamics:} A two-track model with a nonlinear tire model is used. The brake, steering and engine subsystems, which are fitted to track data, model the dynamic response of these systems using delay, PT2 and rate-limiting dynamics. The simulation is performed using an ode45 solver with variable step, limited to a maximum of \SI{1}{ms}. Additionally, we model friction variations and vehicle parameter uncertainty.
    \item \textbf{Predictor:} Estimates the future vehicle state through model forward integration, used as the actual state of the vehicle in the error calculation of the controller.
    \item \textbf{Trajectory Dynamics:} Path matching and reference state interpolation based on an offline-generated trajectory.
\end{itemize}

The following subsections describe modifications to the original control approach which can be classified as robustification and a more general implementation.

\subsubsection{Imposed Error Dynamics}
We made several simplifications for the sake of stability. First, instead of
$\dot{\phi}_{\text{ref}} = \kappa_{\text{ref}} \left( \frac{V \cos \Delta \phi}{1 - \kappa_{\text{ref}}\, e} \right)$, the simplified expression $\dot{\phi}_{\text{ref}} = \kappa_{\text{ref}}\, V$ is used, where $e$ denotes the lateral error, $\Delta \phi$ the course angle error and $\kappa_{\text{ref}}$ the reference curvature. When path tracking errors become too large, the original formulation can introduce instability.

Second, $\dot{r}_{\text{syn}}$  is simplified to $\dot{r}_{\text{syn}}  = \dot{r}_{\text{ref}}$.
In the derivation of Goh \textit{et al.}  there are additional terms that are acknowledged to be negligible \cite[p.3]{Goh.2020}.

\subsubsection{Tire Slip Mapping}
Since no independent wheel speed control is possible,
$\gamma_{\text{RL}} = \gamma_{\text{RR}} = \gamma_{\text{des}}$ cannot be achieved. 
This is not a problem as long as the thrust angle is achieved for the rear axle as a whole. Eq. \ref{eq:TSMmod} defines the computation of the desired wheel speed, where $t$ denotes the track width, and $F_{z,RL}$ and $F_{z,RR}$ represent the vertical loads on the rear left and right tires. Vertical tire loads $F_{z,RL}$ and $F_{z,RR}$ are estimated in real-time using a simple load-transfer model using longitudinal and lateral acceleration ($a_x, a_y$), vehicle mass, track width, and center-of-gravity height. The third term ensures that the thrust angle is achieved even for cases of uneven load distribution. For instance, if $F_{z,RL}$ becomes zero and the entire load of the rear axle is carried by the right-hand tire, the equation becomes equivalent to the formulation of Goh \textit{et al.}  for $\omega_{des,RR}$ \cite[p.6]{Goh.2020}.
\begin{align}
    \omega_{\text{des}} &= \frac{V \cos{\beta}}{R} - \frac{V\sin{\beta} - b \, r}{R \tan{\gamma_{\text{des}}}} \notag \\ &\qquad + \frac{t \, r}{R} \left(\frac{F_{z,\text{RR}}}{F_{z,\text{RL}} + F_{z,\text{RR}}} - 0.5 \right) \label{eq:TSMmod}
\end{align}

\subsubsection{Nonlinear Model Inversion (NMI)}
Our NMI implementation introduces three key modifications to the baseline for improved accuracy and robustness.

First, to better match our vehicle's behavior, we replaced the Fiala-Brush tire model with a more accurate Pacejka MF5 model for computing the tire forces.

\begin{figure}[h]
\centering
\includegraphics[width=1\columnwidth]{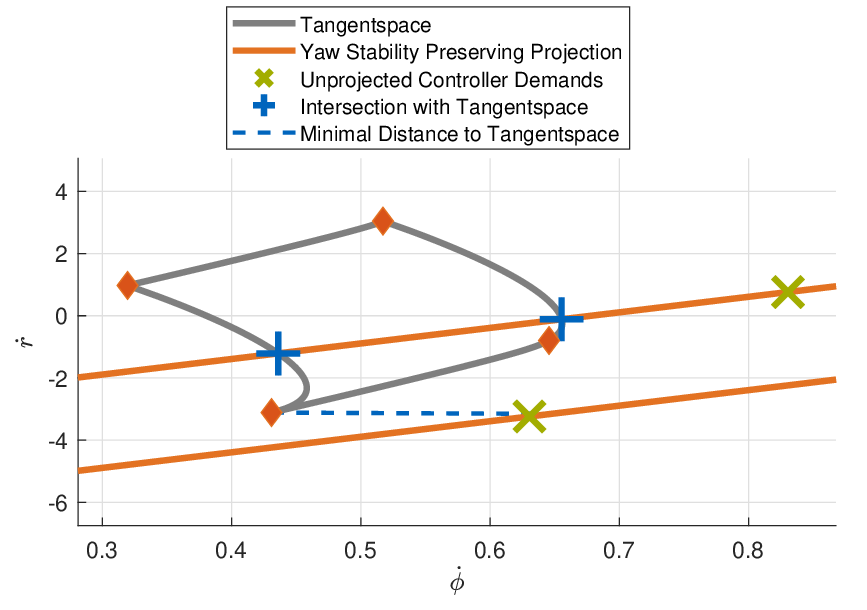}
\caption{Projection of controller demands for two cases, where $r = \SI{0.6}{\radian\per\second}, \beta = \SI{-0.7}{\radian}, V = \SI{13.8}{\meter\per\second}$. The controller demand to the upper right gets projected onto the boundary of the tangent space via the yaw stability preserving projection (YSPP). The lower demand gets calculated via the closest distance of its YSPP to the tangent space.} 
\label{fig:Tangent Space}
\end{figure}

Second, we solve the inversion by formulating an explicit system of nonlinear equations. The system is derived from single-track model equations \cite[p.3]{Goh.2020}, where the rear lateral force $F_{yr}^{\mathrm{I}}$ and the rear longitudinal force $F_{xr}^{\mathrm{III}}$ are expressed as functions of the steering angle $\delta$. The symbols $I_z$ and $m$ denote the yaw inertia and mass; $a$ and $b$ are the distances from the center of gravity to the front and rear axles; $V$ is the velocity; and $\beta$ is the sideslip angle. This formulation yields a single equation in terms of $\delta$ (Eq.~\ref{eq:NMI3}), which is solved using the Newton-Raphson method.
\begin{align}
    F_{yr}^{\mathrm{I}} &= -\frac{I_z \, \dot{r}_{\text{des}} - a \, F_{yf} \cos(\delta)}{b} \label{eq:NMI1}\\
    F_{xr}^{\mathrm{III}} &= \sqrt{\mu_r^2 \, F_{zr}^2 - {F_{yr}^{\mathrm{I}}}^{2}} \label{eq:NMI2}\\
    f(\delta) &= F_{yf} \cos(\beta - \delta) + F_{yr}^{\mathrm{I}} \cos(\beta) \notag \\ &\qquad - F_{xr}^{\mathrm{III}} \sin(\beta) - V \, m \, \dot{\phi}_{\text{des}} \stackrel{!}{=} 0 \label{eq:NMI3}
\end{align}
If the controller's demand for ($\dot{r}_{des}, \dot{\phi}_{des}$) lies outside the feasible tangent space, we project it onto the space's boundary. The point on the boundary closest to the original demand is chosen, ensuring that commands remain physically achievable while prioritizing sideslip stability.
Figure \ref{fig:Tangent Space} visualizes the tangent space for a certain vehicle state and shows the projection for two different controller demands. 

Finally, we introduce a constraint to prevent an intermittently observed drop in sideslip caused by a slow torque response in situations with high rates of $F_{xr,des}$. To mitigate this, we limit the allowable error between the desired and actual wheel speed $\Delta\omega = \omega_{des} - \omega_{act}$, introducing an additional constraint on the rear longitudinal force ($F_{xr,des}$) within the NMI. As a result the steering angle command $\delta_{\text{cmd}}$ adapts to the slow torque and wheel speed response, stabilizing the sideslip dynamics.

\subsection{Predictor for Actuator Latency Compensation}

\begin{figure}
\centering
\includegraphics[width=1.0\columnwidth]{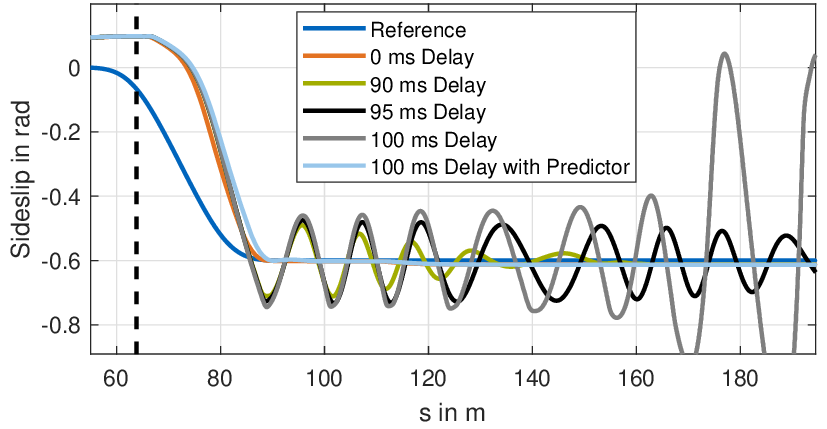}
\caption{Sideslip Tracking Performance with varying synthetic latencies on the actuator signals. With rising delay time oscillations start to form. At over \SI{95}{ms} of actuation latency, the oscillations get catastrophic. The use of the predictor mitigates these effects.}
\label{fig:Oszillation}
\end{figure}

 In contrast to the MARTY vehicle, the internal combustion engine (ICE) and steering system of the test vehicle exhibit significant actuator latencies, as shown in Sec. \ref{subsec:sim}. These delays between a command and its execution can induce oscillations and instability, rendering the original control approach ineffective. An idealized simulation with added actuator latencies highlights this issue (Fig. \ref{fig:Oszillation}), consistent with observations by Joa \textit{et al.}  \cite{Joa.2020}.

To mitigate this, a state predictor module is implemented to compensate for delays inspired by the state space model forecast by Goh \textit{et al.}  \cite{Goh.2024}. The predictor forecasts the vehicle's state forward in time over a prediction horizon set to approximate the mean actuator latency. The drift controller uses these predicted future states for its calculations, effectively aligning its commands with the vehicle's delayed response.

\begin{figure}
    \centering
\begin{tikzpicture} [transform shape, >=stealth', thick, node distance=4cm]

\tikzstyle{block} = [draw, rectangle, minimum height=2em, minimum width=3em]

\node [block, minimum height=6em, text width=5.5em, TUMOrange] (OPred) {$\omega$-Predictor};
\node [block, below of=OPred, node distance=3.3cm, minimum height=11em, text width = 4.2em, yshift=0mm, TUMBlue] (Pred) {Predictor};

\draw [->] ([xshift=-35mm,yshift = 9mm]OPred.west) -- ([yshift = 9mm]OPred.west) node[midway, above] {$\omega^i$};
\draw [->] ([xshift=-35mm,yshift = 3mm]OPred.west) -- ([yshift = 3mm]OPred.west) node[midway, above] {$r^{i-N_{\omega}},\beta^{i-N_{\omega}},V^{i-N_{\omega}},...$};
\draw [->] ([xshift=-35mm,yshift = -3mm]OPred.west) -- ([yshift = -3mm]OPred.west) node[midway, above] {$\tau^{i-N_{\omega}},...$};
\draw [->, TUMBlack] ([xshift=-35mm,yshift = -9mm]OPred.west) -- ([yshift = -9mm]OPred.west) node[midway, above] {$F_{z,\text{RL}}^i,F_{z,\text{RR}}^i...$};

\draw [->] ([xshift=-25mm,yshift = 15mm]Pred.west) -- ([yshift = 15mm]Pred.west) node[midway, above] {$r^i,\beta^i,V^i$};
\draw [->, TUMOrange] ([xshift=-25mm,yshift = 9mm]Pred.west) -- ([yshift = 9mm]Pred.west) node[midway, above] {$\omega_{\text{Present}}$};
\draw [->] ([xshift=-25mm,yshift = 3mm]Pred.west) -- ([yshift = 3mm]Pred.west) node[midway, above] {$x^i,y^i,\phi^i$};
\draw [->] ([xshift=-25mm,yshift = -3mm]Pred.west) -- ([yshift = -3mm]Pred.west) node[midway, above] {$\delta^i, \tau^i$};
\draw [->] ([xshift=-25mm,yshift = -9mm]Pred.west) -- ([yshift = -9mm]Pred.west) node[midway, above] {$\delta_{\text{cmd}}^{i-N_{\delta}},\tau_{\text{cmd}}^{i-N_{\tau}},...$};
\draw [->, TUMBlack] ([xshift=-25mm,yshift = -15mm]Pred.west) -- ([yshift = -15mm]Pred.west) node[midway, above] {$F_{z,\text{FL}}^i,F_{z,\text{FR}}^i...$};

\draw [->, TUMOrange] ([yshift = 0mm]OPred.east) -- ([xshift=15mm,yshift = 0mm]OPred.east) node[midway, above] {$\omega_{\text{Present}}$};

\draw [->, TUMBlue] ([yshift = 9mm]Pred.east) -- ([xshift=20mm,yshift = 9mm]Pred.east) node[midway, above] {$r_P, \beta_P, V_P$};
\draw [->, TUMBlue] ([yshift = 3mm]Pred.east) -- ([xshift=20mm,yshift = 3mm]Pred.east) node[midway, above] {$\omega_P$};
\draw [->, TUMBlue] ([yshift = -3mm]Pred.east) -- ([xshift=20mm,yshift = -3mm]Pred.east) node[midway, above] {$x_P, y_P, \phi_P$};
\draw [->, TUMBlue] ([yshift = -9mm]Pred.east) -- ([xshift=20mm,yshift = -9mm]Pred.east) node[midway, above] {$\delta_P, \tau_P$};

\end{tikzpicture}
    \caption{The predictor structure.}
    \label{fig:Predictor}
\end{figure}
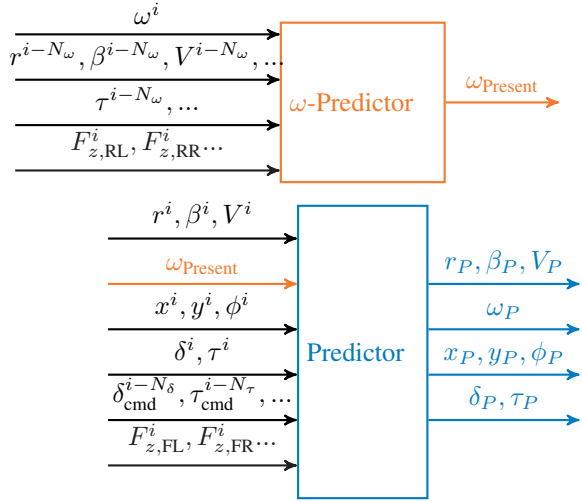

The predictor uses a two-track vehicle dynamics model to propagate the current state. Its inputs are the currently measured vehicle states (e.g. position $x^i, y^i$, velocity $V^i$, yaw rate $r^i$, wheel speed $\omega_{present}$), actuator states ($\delta^i, \tau^i$) and the controller commands ($\delta_{\text{cmd}}$, $\tau_{\text{cmd}}$) from the previous time steps. By applying the previous commands to the current state, the model simulates the physical delay of the system. A pre-filter estimates the present wheel speed from its delayed measurement before this value is passed to the main predictor model. The predictor structure is shown in Fig. \ref{fig:Predictor}.

The dynamics of the predictor are described by the state-space model $\dot{\mathbf{X}} = f_{\text{dyn}}(\mathbf{X}, \mathbf{U})$, where the state vector is $\mathbf{X} = [r, \beta, V, \omega, x, y, \psi, \delta, \tau]^T$ and the input is $\mathbf{U} = [\delta_{\text{cmd}}, \tau_{\text{cmd}}]^T$. The primary state derivatives are given by:
\begin{align}
    \dot{r} &= \frac{1}{I_z} \Bigl( a F_{yf} \cos\delta - b F_{yr} + \frac{t}{2} \sin\delta(F_{y,FL} - F_{y,FR}) \notag \\
    & \qquad + \frac{t}{2} (F_{x,RR} - F_{x,RL}) \Bigr) \label{eq:rdot} \\
    \dot{\beta} &= \frac{1}{mV} \bigl( F_{yf} \cos(\delta - \beta) + F_{yr} \cos\beta - F_{xr} \sin\beta \bigr) - r \label{eq:betadot} \\
    \dot{V} &= \frac{1}{m} \bigl( -F_{yf} \sin(\delta - \beta) + F_{yr} \sin\beta + F_{xr} \cos\beta \bigr) \label{eq:Vdot}
\end{align}
The remaining derivatives describe the wheel speed ($\dot{\omega}$), kinematics ($\dot{x}, \dot{y}, \dot{\psi}$) and actuator responses ($\dot{\delta}, \dot{\tau}$). The state propagation over the prediction horizon is performed using a second-order Runge-Kutta numerical integration with a step size of \SI{10}{ms} over a total horizon of \SI{170}{ms}. This specific horizon was chosen through empirical ablation and yielded the best tracking and oscillation suppression performance. The resulting performance increase due to the predictor is shown in Sec. \ref{subsec:pred_val}.

\subsection{Velocity Control via Supplementary Braking}

During dynamic transitions in the figure-eight maneuver, particularly when the sideslip magnitude is low, the vehicle was observed to accelerate. This unwanted increase in speed leads to increased lateral path tracking errors. To counteract this behavior, a supplementary brake controller was introduced to regulate the vehicle's velocity.

The controller is implemented as a simple proportional controller that becomes active when the vehicle's velocity, $V$, surpasses a predefined target equilibrium speed, $V_{\text{eq}}$. The commanded brake torque, $\tau_{\text{brk,cmd}}$, is calculated according to the following control law:
\begin{equation}
\label{eq:brake_control}
\tau_{\text{brk,cmd}} = -k_{\text{brk}} (V - V_{\text{eq}}) \quad \text{for} \quad V > V_{\text{eq}}
\end{equation}

The brake torque command is applied independently and is not considered within the main nonlinear model inversion of the drift controller. Consequently, braking acts as an external disturbance that the primary controller must reject. This affects the yaw and course angle dynamics and, by introducing a longitudinal force at the front axle, reduces the peak available lateral force. To minimize this interference with the primary objectives of sideslip and path tracking, the commanded brake torque was saturated at a relatively low total maximum value of $2000 \, \text{Nm}$. The effectiveness of this additional controller is shown in Sec. \ref{subsec:vel_control}

\section{RESULTS}
This chapter presents the identification of the vehicle actuator models used in the dynamics model and the predictor. We follow up with a description of our real-world experimental setup and show the effectiveness of our added predictor and velocity controller. We conclude with a comparison of our simulation and real-world experiments and compare our solution with the baseline controller.

\subsection{Simulation Development and Validation}
\label{subsec:sim}
The controller was developed in a MATLAB/Simulink simulation environment using a two-track vehicle model with a MF5 Pacejka tire model and actuator models. The reference maneuver was a symmetric figure-eight drift.
Well-tuned actuator models were necessary to validate the simulation environment for controller development. Initial models proved insufficient, necessitating a data-driven refinement process focused on the steering and the engine.

\begin{figure}
\centering
\includegraphics[width=1\columnwidth]{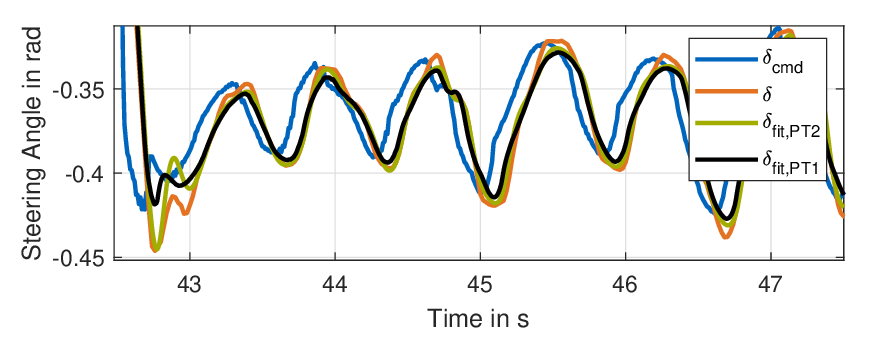}
\caption{Identification of the Steering Model and comparison of a PT1 and PT2 based model.}
\label{fig:SteeringFit}
\end{figure}

A simple first-order (PT1) steering model proved inadequate. During rapidly changing steering inputs, the measured steering angle exhibited a distinct overshoot characteristic, indicative of the inertia of the physical steering rack. The steering was modeled using second-order (PT2) dynamics. The parameters for this PT2 model were identified by fitting its response to the measured data. The initial and final fit is shown in Fig.~\ref{fig:SteeringFit}. The mean peak-to-peak phase delay between the actual steering angle $\delta$ and the steering angle command $\delta_{\mathrm{cmd}}$ is $140 ~\text{ms}$.

Modeling ICE torque delivery is challenging. An initial model with first-order (PT1) dynamics with additional response delay and an engine torque map failed to capture the state-dependent behavior of the engine. Experimental data showed that the rate of torque increase was not constant; it was slow at low torque levels, rose more sharply in a mid-range, and then slowed again at higher output.

\begin{figure}
\centering
\includegraphics[width=1\columnwidth]{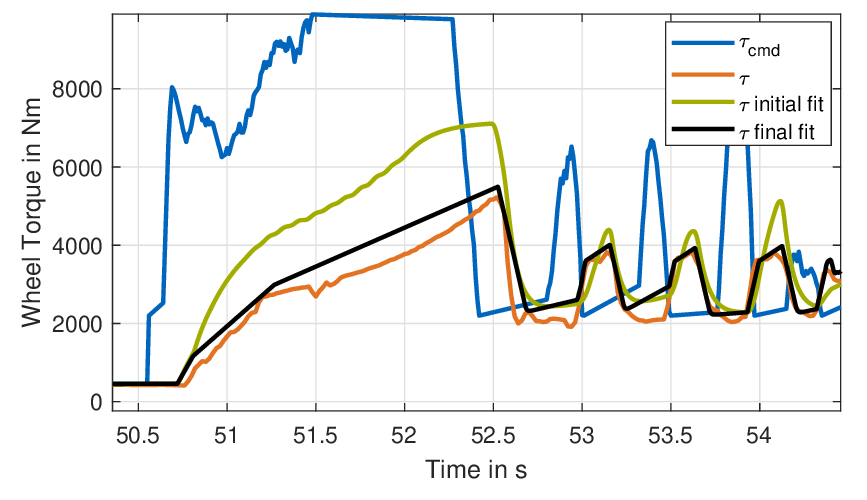}
\caption{Identification of the engine model. A sufficient fit could be achieved with a state-dependent parameter map.}
\label{fig:EngineFit}
\end{figure}

To address this, a state-dependent engine torque model based on real-world data was developed. Instead of a single time constant, the model defines the maximum rate of change of the engine torque output $\tau$ as a function of the current engine torque. This state-dependent approach is shown in Fig~\ref {fig:EngineFit}. The mean peak-to-peak phase delay between the actual engine torque $\tau$ and the commanded engine torque $\tau_{\mathrm{cmd}}$ is $280\,\mathrm{ms}$.

\subsection{Real-World Experiment Setup}
\label{subsec:experiment_setup}
The full control stack was deployed on our Mercedes-AMG GT\,63\,S 4-door test platform with an onboard computer (Vector Informatik VN Control Unit), high-precision localization based on a differential inertial navigation unit (OxTS RT3000 GNSS), and integrated CAN-based actuation. The actuators are Original Equipment Manufacturer (OEM) components and can be accessed via CAN. The vehicle can be set to drift mode, where the active center differential decouples the front axle and the electronic limited-slip differential locks the rear axle. The vehicle weighs \SI{2200}{kg} with a full tank and produces \SI{470}{kW} and \SI{900}{Nm} from its \SI{3982}{\cubic\centi\metre} V8 Biturbo engine. The controller runs at a fixed frequency of 100 Hz.

We conduct experiments on an asphalt surface with mixed conditions, including wet, damp, and dry spots. The vehicle executes figure-eight and circular drift maneuvers. We design the reference trajectories geometrically and set a target slip angle of \SI{0.6}{\radian}. Higher values were possible, but were not reliably achievable above \SI{0.65}{\radian}, because the steering angle command would regularly exceed the steering lock, leading to spin-outs. Transitions between circles follow clothoid curves. The radius of the figure-eight circles is set to \SI{22}{m}, and the centers of the circles are placed \SI{45}{m} apart.

\subsection{Predictor Module}
\label{subsec:pred_val}
\begin{figure}[!ht]
\centering
\includegraphics[width=1.0\columnwidth]{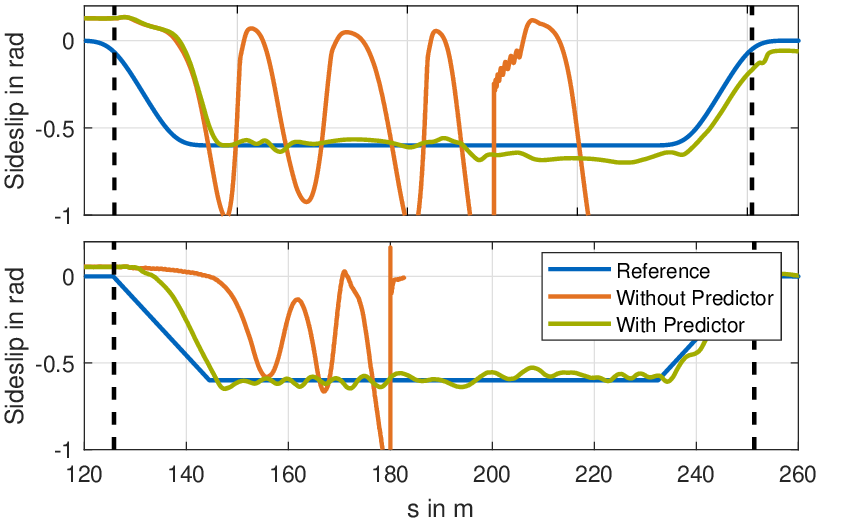}
\caption{Sideslip stabilization with and without predictor. Top: Closed-loop simulation. Bottom: Real vehicle.}
\label{fig:StabilizationWithPredictor}
\end{figure}

Different actuator latencies create phase delays between steering and engine torque. These delays induce divergent oscillations in both simulation and real-world experiments, as shown on a 10\,m radius circular trajectory in Fig. \ref{fig:StabilizationWithPredictor}.

The baseline controller without predictions is unable to maintain stability. Sideslip oscillations grow in amplitude until a spin-out occurs. Analysis of the yaw rate dynamics shows the culprit: the delayed actuator response causes the vehicle to consistently overshoot and undershoot the yaw rate target, feeding the instability. The introduction of the state predictor module successfully resolves this instability in simulation as well as in real-world experiments. By providing the controller with a future state, its commands are effectively timed with the vehicle's delayed response. 

\subsection{Velocity Control Module}
\label{subsec:vel_control}

The velocity controller notably improves lateral tracking and mitigates velocity overshoots, particularly during transitions. Without velocity control, the torque command can reach high values to maintain rear tire saturation during periods of low sideslip, resulting in unwanted longitudinal acceleration as the thrust direction aligns with the trajectory.

\begin{figure}[!ht]
\centering
\includegraphics[width=1.0\columnwidth]{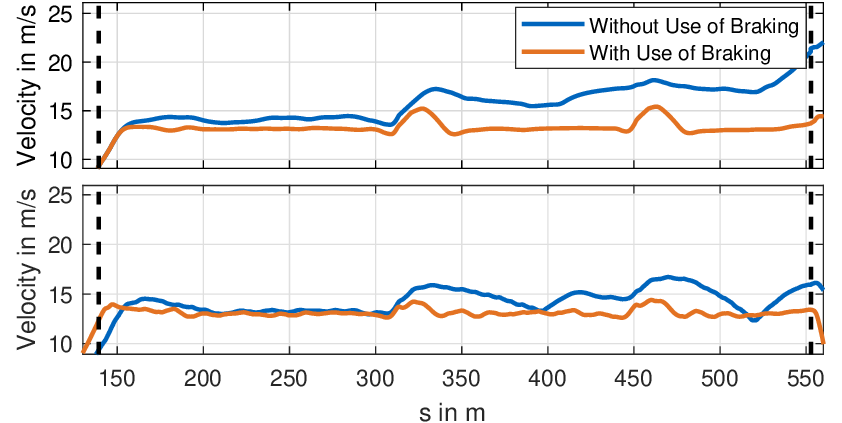}
\caption{Comparison of the velocity with and without the additional velocity controller. Especially in the transitions there is a notable speed increase. Top: Closed-loop simulation. Bottom: Real vehicle.}
\label{fig:SpeedUpIssue}
\end{figure}

\begin{figure}
\centering
\includegraphics[width=1.0\columnwidth]{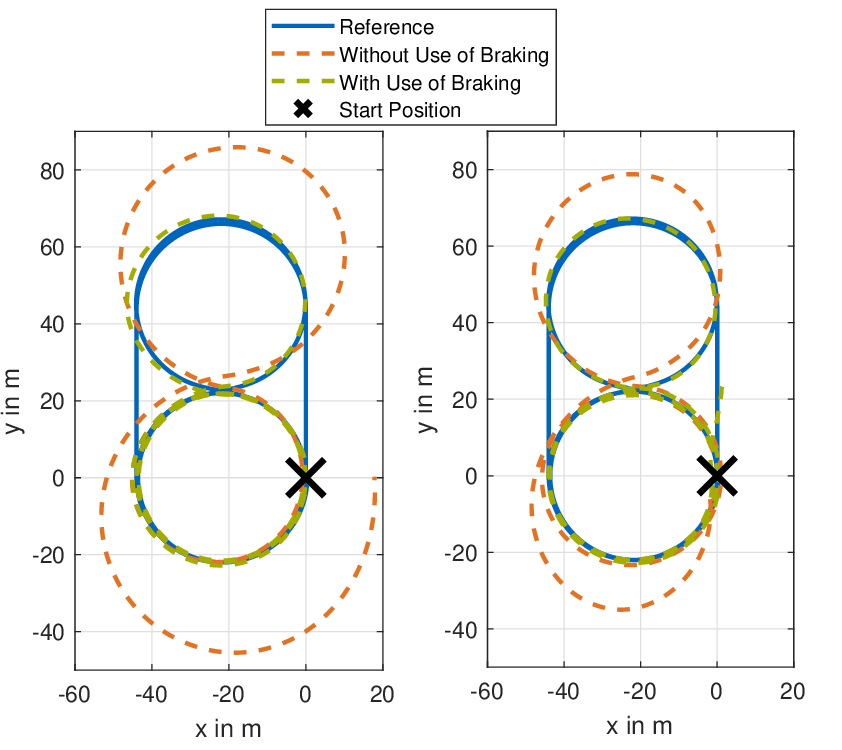}
\caption{Path Tracking Performance with and without Speed Control (Predictor already included). The result of the increased velocity is observable in increased lateral error. Left: Closed-loop simulation. Right: Real vehicle.}
\label{fig:PathtrackingPerformance}
\end{figure}

As a consequence, the vehicle enters the next circle at a higher-than-optimal speed, leading to increased lateral deviations due to the controller’s prioritization of the sideslip target. This speed-up during the transition is shown in Fig. \ref{fig:SpeedUpIssue}.
The incorporation of the velocity controller resolves this issue, and the resulting improvement in velocity overshoots and path-tracking performance can be observed in Fig.~\ref{fig:SpeedUpIssue} \& \ref{fig:PathtrackingPerformance}.

\subsection{Final Results on the Figure-Eight Trajectory}

Our reformulated approach, incorporating the predictor and velocity controller, yielded strong results on our target vehicle. Table~\ref{tab:performance} compares the performance of the adapted baseline approach from Goh \textit{et al.} with our reformulated and extended approach, both in simulation and on the real vehicle. Without our extensions, path tracking was unsuccessful and reliably resulted in spin-outs. By contrast, including the predictor and velocity controller enabled robust tracking of a figure-eight trajectory. Experiments repeated on the real vehicle confirmed these results and even outperformed the simulation. This improvement is largely due to the simulation model being intentionally parameterized to be more challenging, with several parameters set to worst-case values within their plausible range. This approach created a demanding simulation environment that reproduced many on-track effects, facilitating quick iterations in our development cycles.

\begin{table}
\caption{Evaluation of the baseline controller against our extended approach including predictor and velocity control in our closed-loop simulation, using the identified AMG vehicle model. The third column shows the performance of our our extended approach on the real vehicle for comparison. The control parametrization has not been changed between column two and three.}
\centering
\begin{tabular}{lccc}
\toprule
Metric & Sim (Goh) & Sim (Ours) & Real-world (Ours) \\
\midrule
Max Sideslip & $\geq\SI{1.5}{\radian}$ &$\SI{0.64}{\radian}$ & $\SI{0.66}{\radian}$ \\
Max Lateral Error & $\geq\SI{10.0}{m}$ &$\SI{2.4}{m}$ & $\SI{1.1}{m}$ \\
Max Heading Error & $\geq\SI{1.5}{\radian}$ & $\SI{0.41}{\radian}$ & $\SI{0.1}{\radian}$ \\
\bottomrule
\end{tabular}
\label{tab:performance}
\end{table}

\begin{figure}
\centering
\includegraphics[width=1.0\columnwidth]{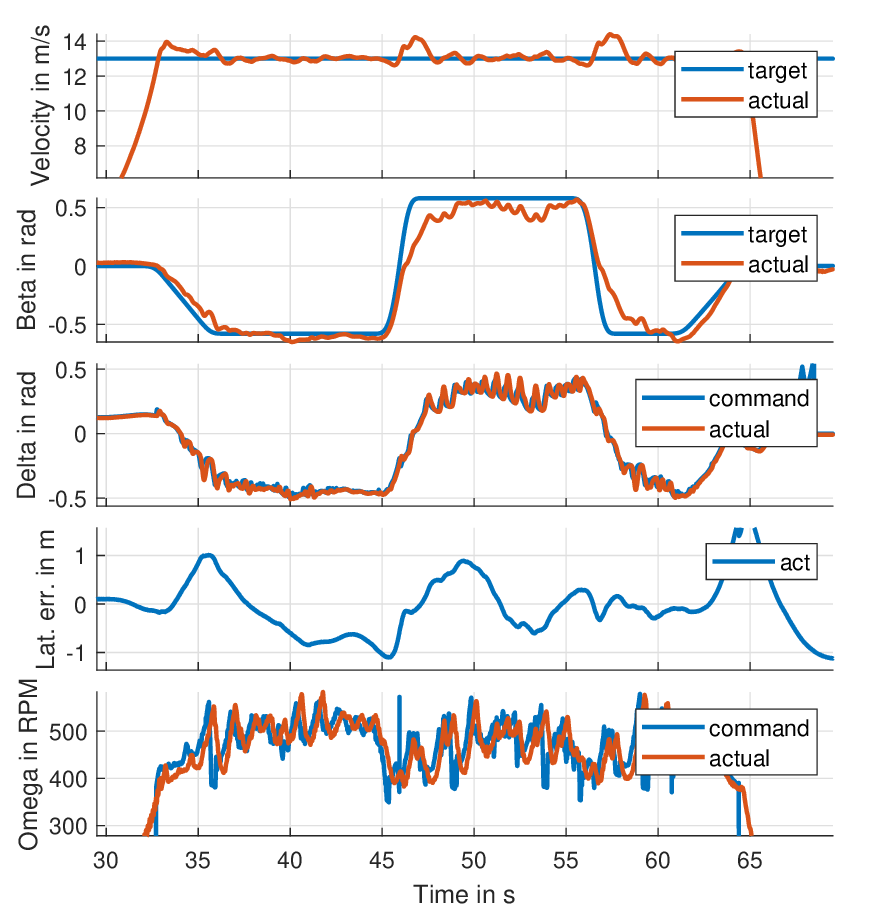}
\caption{Final figure-eight performance on the real Mercedes-AMG GT\,63\,S vehicle.}
\label{fig:final_eval}
\end{figure}

The performance on the full figure-eight on the real vehicle is shown in Fig. \ref{fig:final_eval}. We compare our performance against the data from Goh \textit{et al.} on the MARTY vehicle platform \cite{Goh.2019} in Tab. \ref{tab:performance2}. Overall we conclude that despite the more challenging vehicle platform we could achieve similar overall performance, even improving the performance in certain metrics like lateral deviation or sideslip overshoot. Also, our velocity variation is slightly narrower (\SI{3.0}{\meter\per\second} vs. \SI{1.8}{\meter\per\second}).
The peaks in the lateral deviation and velocity overshoots occurred during the entry, exit, and transition phases of the figure-eight, which are the most difficult to handle. However there are slight differences in our experimental setup. Due to our lower mechanical steering angle lock at \SI{34}{\degree} compared to their \SI{38}{\degree}, we set a slightly lower slip angle target and increased the corner radius. However, we do not expect those differences to invalidate the comparison. 

\begin{table}
\caption{Comparison of the real-world results achieved by Goh \textit{et al.}  \cite{Goh.2019} on the MARTY vehicle against our results on the AMG vehicle.}
\centering
\begin{tabular}{lcc}
\toprule
Metric & Goh \textit{et al.}  \cite{Goh.2019} & \textbf{Ours }\\
\midrule
Single circle radius & $\SI{18}{\meter}$ & $\SI{22}{\meter}$ \\
Max Sideslip Target  & $\SI{0.7}{\radian}$ & $\SI{0.6}{\radian}$ \\
Max Sideslip Overshoot & $\SI{0.08}{\radian}$ & \textbf{$\SI{0.06}{\radian}$} \\
Max Abs. Lateral Error & $\SI{1.4}{m}$ & \textbf{$\SI{1.1}{m}$} \\
Max Abs. Heading Error & $\SI{0.11}{\radian}$ & \textbf{$\SI{0.1}{\radian}$} \\
Max Abs. Yaw Rate & $\SI{2}{\radian\per\second}$ & \textbf{\SI{0.95}{\radian\per\second}} \\
Min/Max Velocity & \SI{10.8}{} / \SI{13.8}{\meter\per\second} & \SI{12.6}{} / \SI{14.4}{\meter\per\second} \\
\bottomrule
\end{tabular}
\label{tab:performance2}
\end{table}

\section{DISCUSSION \& CONCLUSION}

Our findings highlight that latency compensation is critical for applying drift control to production vehicles. While prior drift controllers perform well under ideal actuation, real-world ICE systems demand predictive behavior to maintain stability.

The predictor module was the most impactful enhancement. Its integration enables the system to anticipate the future vehicle response and maintain phase coherence between model and real dynamics. Our approach extends the original framework to retain drift stability on platforms with significant latency. Finally, on the AMG vehicle, our controller successfully replicated and even slightly improved the tracking performance that the baseline framework demonstrated on the MARTY platform, despite the additional challenges posed by actuator latency. In addition, we were able to avoid the need for independent wheel speed control on the rear axle. Although a nonlinear MPC could theoretically handle prediction and control in a unified framework, our modular approach is computationally lightweight and more easily interpretable. This makes it suitable for deployment on existing automotive-grade hardware without the model tuning and computational overhead associated with MPC. Unlike learning-based systems, our architecture remains interpretable and does not require large-scale training data.

Although our empirical results validate the effectiveness of the framework, several limitations remain for future work. First, unlike linear delay compensation methods (e.g., Smith predictors), our forward-integration predictor handles the non-linearities of tire limits but lacks a formal mathematical proof of closed-loop stability. Second, our real-world testing was conducted on structured geometric trajectories (circles and figure-eights). Future research will evaluate the controller's robustness on arbitrary, varying-curvature tracks. Furthermore, we note that tire model fidelity and friction estimation still limit performance. Sensor-based adaptation or online learning could further improve robustness \cite{GohTireModels}.

\section*{Acknowledgements}

The authors thank Mercedes-AMG for their support and collaboration in the project. In particular, the authors acknowledge Florian Pejak and Szabolcs Dora for their contributions through discussions and technical support. The authors thank Vuk Boskovic, who supported this work through his Master Thesis.
This work was supported by the Munich Institute of Robotics and Machine Intelligence (MIRMI).



\bibliographystyle{IEEEtran}
\bibliography{Bibliography}

\end{document}